\documentclass{article}
\usepackage{arxiv}
\usepackage{natbib}
\setcitestyle{authoryear,round,citesep={;},aysep={,},yysep={;}}

\usepackage{xcolor}
\usepackage[utf8]{inputenc} 
\usepackage[T1]{fontenc}    
\usepackage{hyperref}       
\usepackage{url}            
\usepackage{booktabs}       
\usepackage{amsfonts}       
\usepackage{nicefrac}       
\usepackage{algorithmic}
\usepackage{microtype}      
\usepackage{lipsum}		
\usepackage{wrapfig}
\usepackage{fancybox}
\usepackage{graphicx}
\usepackage[boxruled]{algorithm2e}
\usepackage{enumitem}
\usepackage{float}

\usepackage{doi}
\usepackage{pgfplots}
\usepackage{pgf}
\usepackage{calc}  
\usepackage{enumitem}  
\usepackage{geometry}
\usepackage{mathtools}

\usepackage{bbm}
\usepackage{caption}
\usepackage{subcaption}
\usepackage{scalerel}

\usepackage{tikz}
\usepackage{adjustbox}

\DeclarePairedDelimiter\floor{\lfloor}{\rfloor}

\newtheorem{assumption}{Assumption}
{\begin{Sbox}\begin{minipage}}%
{\end{minipage}\end{Sbox}\fbox{\TheSbox}}

\newcommand{\quotes}[1]{``#1''}

\definecolor{YellowOrange}{RGB}{247, 146, 28}
\definecolor{NavyBlue}{RGB}{2, 110, 184}
\definecolor{DarkGreen}{RGB}{85, 168, 104}
\definecolor{DarkRed}{RGB}{196, 78, 82}

\date{}

\hypersetup{
	pdftitle={Tuning Confidence Bound for Stochastic Bandits with Bandit Distance},
	pdfsubject={statistics},
	pdfauthor={Xinyu Zhang},
	pdfkeywords={Multi-Armed Bandits, UCB, Confidence Bound},
	}

\title{Tuning Confidence Bound for Stochastic Bandits with Bandit Distance}

\author{Xinyu Zhang\\
	UC San Diego\\
	\texttt{xiz368@eng.ucsd.edu}
	\And
	Srinjoy Das\\
	West Virginia University\\
	\texttt{srinjoy.das@mail.wvu.edu} \\
	\And
	Ken Kreutz-Delgado\\
	UC San Diego\\
	\texttt{kreutz@eng.ucsd.edu} 
}

\begin{document}
\maketitle

\begin{abstract}
	We propose a novel modification of the standard upper confidence bound (UCB)
	method for the stochastic multi-armed bandit (MAB) problem which tunes the
	confidence bound of a given bandit based on its distance to others. Our UCB
	distance tuning (UCB-DT) formulation enables improved performance as
	measured by expected regret by preventing the MAB algorithm from focusing on
	non-optimal bandits which is a well-known deficiency of standard UCB.
	"Distance tuning" of the standard UCB is done using a proposed distance
	measure, which we call bandit distance, that is parameterizable and which
	therefore can be optimized to control the transition rate from exploration
	to exploitation based on problem requirements. We empirically demonstrate
	increased performance of UCB-DT versus many existing state-of-the-art
	methods which use the UCB formulation for the MAB problem. Our contribution
	also includes the development of a conceptual tool called the
	\textit{Exploration Bargain Point} which gives insights into the tradeoffs
	between exploration and exploitation. We argue that the Exploration Bargain
	Point provides an intuitive perspective  that is useful for comparatively
	analyzing the performance of UCB-based methods.

\end{abstract}

\section{Introduction}

Multi-armed bandit (MAB) \citep{slivkins_tutorial} can model a broad
range of applications, such as selecting the best website layout for users, or
choosing the most profitable stocks among many. Stochastic bandits is an
important setting in which MAB problems have been studied extensively. One of
the most influential and widely used stochastic bandit policy is the upper
confidence bound method (UCB) \citep{auer2002finite}. UCB works by maintaining a
mean estimation and confidence radius\footnote{We will refer confidence radius
as confidence bound in the rest of the paper.} for each bandit, and selects the
bandit whose sum of mean and confidence bound is the maximum among all bandits
at each step. The confidence bound can grow larger for less frequently used
bandits which represents a higher degree of uncertainty to serve the exploration
purpose.


However, the original UCB algorithm by its nature can lead to unsatisfactory
results by being over-optimistic on non-optimal bandits. In this work, we
propose UCB-DT (Upper Confidence Bound - Distance Tuning), a simple modification
to the original UCB method which makes the confidence bound of a given bandit
depend on its distance to others. Since the UCB-DT policy will select the
largest bandit more often, exploration will naturally lean towards the neighbors
of the largest bandit and prevent the algorithm from focusing on bandits that
are farther away. Our proposed bandit distance is parmeterizable thereby
offering the opportunity of customization over policies through different
distances. Therefore, our formulation can represent a family of policies which
inherently provide the flexibility of both pro-exploration policies, such as
UCB, and pro-exploitation policies like $\epsilon$-greedy.

Moreover, unlike previous UCB-based methods which focus on the $\log$ function
in the confidence bound because of its analytical tractability, our method works
differently by extending the denominator term. Using this enhancement to
standard UCB, we propose a concept named \textit{Exploration Bargain Point} to
provide a novel perspective on analyzing performance of UCB-based methods. Using
our new analysis tool, we intuitively and even graphically show that our method
can always perform better than standard UCB.


We review existing work on the design of confidence bound for UCB in
Sec.~\ref{sec:relate}. While maintaining connections to some of these previous
approaches, we make the following novel contributions in this paper.

\begin{itemize}
	\item We propose UCB-DT policy, which tunes confidence bound by
	      bandit distance. We conduct analysis and numerical experiments to show
	     that our
	      formulation is simple, extensible, and performant.
	      	      	      	      	      	      	      	      	      	      	      	      	      	      	      	      	      	      	      	      	      	      	      	      	      	      	      	      	      	      	      	      	      	      	      	      	      	      	      	      	      	      	      	      	      	      	      	      	      	      	      	      	      	      	      	      	      	      	      	      	      	      	      	      	      	      	      	      	      	      	      	      	      	      	      	      	      	      	      	      	      	      	      	      	      	      	      	      	      	      	      	      	      	      	      	      	      	      	      	      	      	      	      	      	      	      	      	      	      	      	      	      	      	      	      	      	      	      	      	      	      	      	      	      	      	      	      	      	      	      	      	      	      	      	      	      	      	      	      	      	      	      	      	      	      	      	      	      	      	      	      	      	      	      	      	      	      	      	      	      	      	      	      	      	      	      	      	      	      	      	      	      	      	      	      	      	      	      	      	      	      	      	      	      	      	      	      	      	      	      	      	      	      	      	      	      	      	      	      
	\item We present a concept called \textit{Exploration Bargain Point},
	      which provides a novel viewpoint on analyzing performance of upper
	      confidence bound methods.

	      	      	      	      	      	      	      	      	      	      	      	      	      	      	      	      	      	      	      	      	      	      	      	      	      	      	      	      	      	      	      	      	      	      	      	      	      	      	      	      	      	      	      	      	      	      	      	      	      	      	      	      	      	      	      	      	      	      	      	      	      	      	      	      	      	      	      	      	      	      	      	      	      	      	      	      	      	      	      	      	      	      	      	      	      	      	      	      	      	      	      	      	      	      	      	      	      	      	      	      	      	      	      	      	      	      	      	      	      	      	      	      	      	      	      	      	      	      	      	      	      	      	      	      	      	      	      	      	      	      	      	      	      	      	      	      	      	      	      	      	      	      	      	      	      	      	      	      	      	      	      	      	      	      	      	      	      	      	      	      	      	      	      	      	      	      	      	      	      	      	      	      	      	      	      	      	      	      	      	      	      	      	      	      	      	      	      	      	      	      	      	      	      	      	      	      	      	      	      
\end{itemize}

\section{Preliminaries}

Let $k \in \mathbb{Z}^+$ be the number of bandits, $T$ denote the time horizon,
$\mu_i \in \mathbb{R}$ be the unknown mean for the subgaussian reward distribution of bandit $i$,
$\mu_* = \max_i \mu_i$ be the mean reward of the optimal bandit, $B_i$ be the shorthand
for \quotes{$i$th bandit} where $i\in[k]$. In each round $t \in [T] = \{1,2,...,T\}$, the policy
chooses a bandit $A_t \in [k]$ and whose reward is denoted by random variable $X_t$.

We use $\Delta_i = \mu_* - \mu_i$ to represent the suboptimality gap for $B_i$,
and $N_i(t)$ to represent the number of times bandit $i$ gets chosen till $t$.
The regret over $T$ rounds is

\vspace*{-1em}
{
	\small
\begin{equation}
	\mathcal{R}_{T} =\sum_{i=1}^k \Delta_{i} \mathbb{E}\left[N_{i}(T)\right], \text{ where } N_{i}(t) =\sum_{s=1}^{t} \mathbb{I}\left[A_{s}=i\right]
	\label{eq:regret}
\end{equation}
}
\vspace*{-0.5em}

which serves as the main metric for stochastic bandit policies. A policy is
called \textit{asymptotically optimal} if

\vspace*{-1em}

{
	\small
\begin{equation}
\lim _{T \rightarrow \infty} \frac{\mathcal{R}_{T}}{\log (T)}=\sum_{i: \Delta_{i}>0} \frac{2}{\Delta_{i}}
\label{eq:regret_bound}
\end{equation}
\vspace*{-1em}
}

\citep{lai1985asymptotically,burnetas1997optimal} show that the above forms a
regret upper bound for all consistent policies. A policy is called
\textit{sub-UCB} \citep{lattimore2018refining}, which is a stricter requirement
than \textit{asymptotically optimality}, if there exists universal constants
$C_1, C_2 > 0$, such that the regret can be finitely bounded as

\vspace*{-2em}
{
\small
\begin{equation}
\mathcal{R}_{t} \leq C_{1} \sum_{i=1}^{k} \Delta_{i} + C_{2} \sum_{i: \Delta_{i}(\mu)>0} \frac{\log (n)}{\Delta_{i}}
\label{eq:regret_bound_finite}
\end{equation}
}
\vspace*{-1em}

The UCB policy works by choosing bandit $A_t$ such that

\vspace*{-1em}
{
\small
\begin{equation}
A_t = \arg\max_{i \in [k]} \hat{\mu}_i(t-1) + \sqrt{\frac{2 \log (t-1)}{N_i(t-1)}}
\label{eq:ucb}
\end{equation}
}

\vspace*{-.5em}

where the first term $\hat{\mu}_i(t) = (\sum_{c=1}^t \mathbb{I}(A_c = i) X_c) /
N_i(t)$ represents the estimation of $\mu_i$ at time $t$, and the second term
$\sqrt{2 \log (t) / N_i(t)}$ denotes the confidence bound. UCB satisfies
\textit{sub-UCB} requirement \citep{auer2002finite} and is an
\textit{anytime} policy i.e. selection of the optimal bandit in
Eq.~\ref{eq:ucb} does not depend on $T$.

\section{Related Work}
\label{sec:relate}

\textbf{Multi-armed bandits.} MAB \citep{slivkins_tutorial} is a simple yet
powerful framework for decision making under uncertainty. There are many
categories of MAB problems including stochastic bandits proposed by
\citep{gittins1979bandit, lai1985asymptotically, katehakis1987multi}, as the
most classical one, where it is assumed that $\forall t, r_{ti}$ are samples
drawn from a stationary sub-gaussian distribution bound to bandit $i$. Here
$r_{ti}$ represents the reward at time $t$ on arm $i$. Adversarial bandits
\citep{auer1995gambling} assumes that for $\forall t, r_{ti}$ do not have to
belong to any stationary distribution and can be set by an adversary. It is
common to model $r_{ti}$ as a secret codebook set by an enemy who knows the
policy before playing. Contextual bandits \citep{langford2007epoch} introduces
an observable context variable and assumes $r_{ti}$ is drawn a distribution
parameterized by both bandit $i$ and a context variable. There are many other
variants under the MAB framework, whose details are beyond the scope of this
paper. In this paper, we focus on the stochastic bandits problem.

\textbf{UCB formulation for stochastic bandits.} Since being first proposed, UCB
\citep{auer2002finite} has received strong research interest and several
variants of the original UCB policy have been proposed. For example, KL-UCB
\citep{garivier2011kl, cappe2013kullback} and KL-UCB++ \citep{menard2017minimax}
transform the UCB policy as a procedure that calculates the best possible arm
using a Kullback-Leibler divergence bound at each time step. UCBV-Tune
\citep{audibert2009exploration} incorporates the estimated variance of reward
distribution instead of assuming unit variance.

In the context of this paper, several previous authors have proposed
formulations which redesign the confidence bound of standard UCB as shown in the
term of Eq.~\ref{eq:ucb}. MOSS \citep{audibert2010regret} makes the confidence
bound depend on the number of plays for each bandit by replacing $\log(t)$ with
$\log(t / N_i(t))$ in Eq.~\ref{eq:ucb}, and policies similar to MOSS include
OCUCB \citep{lattimore2016regret} and UCB* \citep{garivier2016explore}.
UCB$\dagger$ \citep{lattimore2018refining} improves upon the previous ones
significantly by designing a more advanced log function component.

Compared with these previous approaches, our method works differently by
extending the denominator term rather than the $\log$ function. It turns out to
be intuitive in foresight and delivers strong performance. Moreover, instead of
being non-parametric like most of the above methods, our policy has parameters
which are tunable, which allows that the formulation of our method to encompass
a family of policies.

\section{Method}

\subsection{Intuition}
\label{sec:intuition}

The core idea of UCB-DT is that when a bandit is selected, instead of increasing
the confidence bounds of all other bandits uniformly, we increase them more for
bandits which are similar to the current chosen bandit and vice versa. The
intuition is that a similar bandit has a higher chance to be equally good as the
current chosen one. As a result, this strategy will naturally lean towards the
optimal bandit and spend exploration budget on similarly good bandits and save
unnecessary trials with poor bandits.

To realize this idea, we start by looking at the confidence bound term $\sqrt{2
\log (t) / N_i(t)}$ for $B_i$ in Eq.~\ref{eq:ucb}. Suppose there exists a
distance measure $d(i, j)$ ranges between $[0, 1]$ that compares the distance
between $B_i$ and $B_j$. Then we find that the above idea can be implemented by
replacing $N_i(t)$ as

\begin{equation}
	\begin{split}
		N_i(t) \Rightarrow  N_i(t) + \sum_{j \in [k], j \neq i} d(i, j) N_j(t)
	\end{split}
	\label{eq:Ntilde}
\end{equation}

This modification can shrink the confidence bounds of bandits which are
distant from others, while maintaining the confidence bounds
of those ones that are closer for further exploration.
Furthermore, this modification elegantly depicts the poles of exploration
and exploitation as below:

\begin{description}
	\item[Exploration] When $d(i, j) \equiv 0$, Eq. \ref{eq:Ntilde} will
	degrade to the vanilla UCB case, which is an "optimistic" policy and
	encourages exploration. 
	\item[Exploitation] When $d(i, j) \equiv 1$, Eq. \ref{eq:Ntilde} will degrade to greedy case because $N_i(t) = t$ and $\log (t) /
	t$ can rapidly converge to 0, which purely exploits.
\end{description}

\begin{wrapfigure}[12]{R}{0.57\textwidth}
	\centering
	\begin{minipage}{0.9\linewidth}
		\begin{algorithm}[H]
			\SetKwInOut{Input}{ input}
			\Input{distance measure $d_t$}
			\vspace*{-1em}
			\begin{equation*}
				\begin{split}
					A_t = \arg\max_{i\in[k]} \hat{\mu}_i(t-1) + \sqrt{\frac{2 \log (t-1)}{ \widetilde{N}_i(t-1) }} \\
					\text{where } \widetilde{N}_i(t) = N_i(t) + \sum_{j \in [k], j \neq i} d_t(i, j)  N_j(t)
				\end{split}
			\end{equation*}
			\caption{UCB distance tuning (UCB-DT)}
			\label{alg:ucbdt}
		\end{algorithm}
	\end{minipage}
\end{wrapfigure}

These two special cases correspond exactly to the simple and commonly recognized
truth: \textit{In the stochastic bandit problem, you shall explore just
enough to find the right bandit, then exploit that as long as you can.}


In our formulation, we can model this transition from exploration to
exploitation by customizing $d$. It indicates that if we could find a transition
from $d \equiv 0$ to $d \equiv 1$, like expanding $d$ from the origin to a unit
circle, then we can realize the above truth under the framework in
Eq.~\ref{eq:Ntilde}.

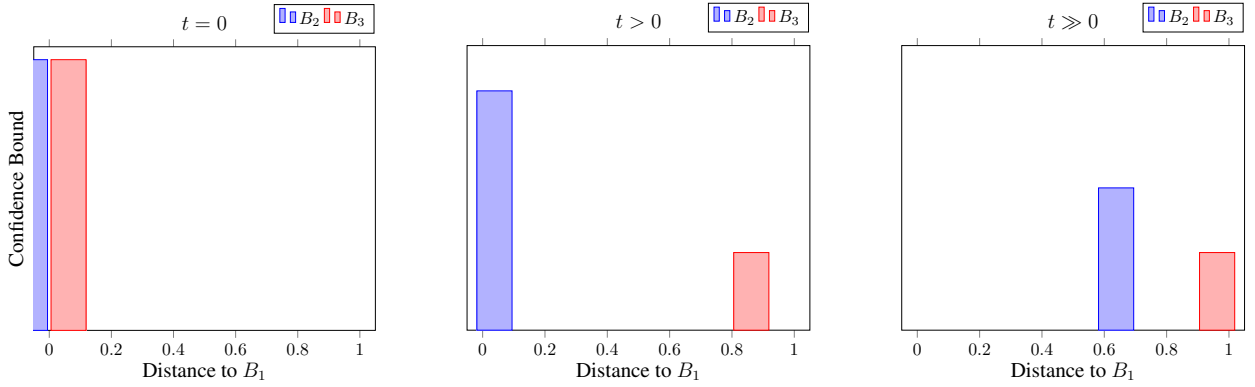
\begin{figure}
	\centering
	\begin{subfigure}[b]{0.3\textwidth}
		\begin{adjustbox}{width=\linewidth}
			\begin{tikzpicture}
				\begin{axis}[
						ymajorticks=false,
						ylabel={\large Confidence Bound},
						ylabel style={yshift=-2.5em},
						enlargelimits=0.05,
						legend style={at={(0.85,1.15)},
							anchor=north,legend columns=-1},
						x label style={at={(axis description cs:0.5,0.0)},anchor=north},
						xlabel={\large Distance to $B_1$},
						ybar,
						bar width=2em,
						title = {\large $t=0$},
						xmin=0,
						xmax=1,
						ymin=0.2,
						ymax=1
					]
					\addplot 
					coordinates {(0,1)};
					\addplot 
					coordinates { (0,1)};
					\legend{$B_2$,$B_3$}
				\end{axis}
			\end{tikzpicture}
		\end{adjustbox} 
		\caption{Given no prior knowledge, the policy assumes $B_1,B_2,B_3$ have no difference,
		 $d_t(1, 2) = d_t(1, 3) = 0$, which renders large confidence bounds and promote 
		exploration.}
	\end{subfigure}
	\hfill
	\begin{subfigure}[b]{0.3\textwidth}
		\begin{adjustbox}{width=\linewidth}
			\begin{tikzpicture}
				\begin{axis}[
						ymajorticks=false,
						ylabel={\phantom{1}},
						ylabel style={yshift=-2.5em},
						enlargelimits=0.05,
						legend style={at={(0.85,1.15)},
							anchor=north,legend columns=-1},
						x label style={at={(axis description cs:0.5,0.0)},anchor=north},
						xlabel={\large Distance to $B_1$},
						ybar,
						bar width=2em,
						title = {\large $t>0$},
						xmin=0,
						xmax=1,
						ymin=0.2,
						ymax=1
					]
					\addplot 
					coordinates {(0.1,0.9)};
					\addplot 
					coordinates { (0.8,0.4)};
					\legend{$B_2$,$B_3$}
				\end{axis}
			\end{tikzpicture}
		\end{adjustbox} 
		\caption{Learning that $B_3$ is far from $B_1$, $d_t(1,3)$ will be close
		to 1, which inhibits its confidence bound to grow larger. Thus, the
		policy will focus on comparing $B_1$ and $B_2$.}
	\end{subfigure}
	\hfill
	\begin{subfigure}[b]{0.3\textwidth}
		\begin{adjustbox}{width=\linewidth}
			\begin{tikzpicture}
				\begin{axis}[
						ymajorticks=false,
						ylabel={\phantom{1}},
						ylabel style={yshift=-2.5em},
						enlargelimits=0.05,
						legend style={at={(0.85,1.15)},
							anchor=north,legend columns=-1},
						x label style={at={(axis description cs:0.5,0.0)},anchor=north},
						xlabel={\large Distance to $B_1$},
						ybar,
						bar width=2em,
						title = {\large $t \gg 0$},
						xmin=0,
						xmax=1,
						ymin=0.2,
						ymax=1
					]
					\addplot 
					coordinates {(0.7,0.6)};
					\addplot 
					coordinates { (0.9,0.4)};
					\legend{$B_2$,$B_3$}
				\end{axis}
			\end{tikzpicture}
		\end{adjustbox} 
		\caption{Given the expanding property, $d_t(1,2)$ will grow
		larger even if $B_1,B_2$ are close. So the confidence bound of $B_2$
		will cease to dominate eventually, policy can focus on $B_1$.}
	\end{subfigure}
	\caption{A visual demonstration for our intuition with an example
		$B_1,B_2,B_3$ whose $\mu_1 > \mu_2 \gg \mu_3$ and $\mu_1 = \mu_2 +
		\epsilon$.}
																																					 
	\label{fig:visual_example}
\end{figure}

Therefore, we write $d$ as time dependent as $d_t$ and condense the above
findings in Alg. \ref{alg:ucbdt}. An example is provided to demonstrate our
ideas in Fig. \ref{fig:visual_example}. A specific instance of $d$ will be
introduced in Sec. \ref{sec:bdistance}.


\subsection{Bandit Distance}

\label{sec:bdistance}

We design the following distance which composes UCB-DT($\mu$).

\begin{description}
	\item[UCB-DT($\mu$)] 
	\begin{equation}	
	d_t(i, j) =|\hat{\mu}_i(t) - \hat{\mu}_j(t)|^{1 / \floor{\gamma N_i(t)}}
	\label{eq:ucbdt}
	\end{equation}
\end{description}

First, it directly measures the distance between two bandits. Second, the more
often a bandit gets pulled, the closer its distances from all other bandits will
approach the maximum value of 1, which means that the policy for this bandit
transitions deeper into exploitation from exploration. Here $\gamma$ is a speed
parameter to control the transition rate. It could be pointed that the
UCB-DT($\mu$) will not work properly for bandits whose $\mu_i \gg 1$ because the
distance will saturate. But we argue that it is common practice to use
normalization to bypass this constraint. Thus, we keep the above design for
simplicity.

\begin{figure}[H]
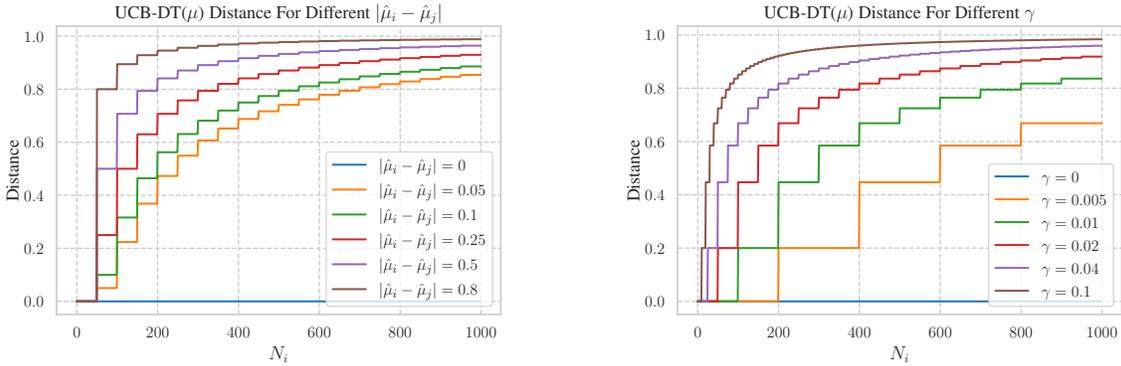

	\centering
	\begin{minipage}{.5\textwidth}
		\centering
		\scalebox{0.5}{
			\input{plots/pgf/bandit-distance-1.pgf}
		}
	\end{minipage}%
	\begin{minipage}{.5\textwidth}
		\centering
		\scalebox{0.5}{
			\input{plots/pgf/bandit-distance-2.pgf}
		}
	\end{minipage}
	\caption{Visualization of the Distance in Eq.~\ref{eq:ucbdt}. $\gamma$ is
	set to $0.02$ in the left figure, and $|\hat{\mu}_i - \hat{\mu}_j|$ is set
	to $0.2$ in the right figure. It can be clearly seen that the greater the
	difference between the mean rewards of two bandits is, the faster that the distance is expanded
	from $0$ to $1$. The rate of convergence of $d$ to $1$ can also be
	controlled by increasing $\gamma$. The curves are jagged because of the
	floor operation $\floor{\gamma N_i}$.}
	\label{fig:bandit_dist_vis}
\end{figure}

In Appendix \ref{apn:more_d}, we provide more designs for $d$ and analyze their
characteristics. In Fig.~\ref{fig:bandit_dist_vis}, we visualize the distance
versus $N_i$ using different $\gamma$ and $|\hat{\mu}_i - \hat{\mu}_j|$.

\subsection{Under Exploration Analysis}
\label{sec:under_exploration}

As compared to standard UCB, our proposed formulation UCB-DT performs less
exploration. In this section, we conduct an analysis based on a novel concept
called \textit{Exploration Bargain Point} to show that our method can always
give better performance than UCB. Based on our analysis, we also provide
practical guidelines on how to set the parameter $\gamma$.


In the following discussion, we assume a scenario of two bandits where $\mu_1 >
\mu_2$, and we also assume that $N_1 \ge N_2$. Since we only have two bandits,
we can regard $N_2(T)$ as the exploration budget spent till time $T$, and $T =
N_1 + N_2$. We conduct our analysis by hindsight\footnote{\quotes{By hindsight}
means we look back at a policy's decisions from time $T$.} in the context of
standard UCB. 

\subsubsection{Exploration Full Point}

If we want to explore enough to ensure $P(A_T = 1) \ge 1 - \delta$, this implies
that we recognize the optimal bandit as the dominant choice. Therefore, based on
Eq.~\ref{eq:ucb_dawn} which is discussed in the formulation of standard UCB in
Appendix \ref{apn:ucb}, we have:

\begin{equation}
	P\left(\sqrt{\frac{2}{n} \log \left(\frac{1}{\delta}\right)} \le \Delta_2 / 2\right) \le
	\delta
	\label{eq:full_inequality}
\end{equation}

where $\sqrt{\frac{2}{n} \log \left(\frac{1}{\delta}\right)}$ is the largest
possible deviation of mean estimation. Since we have $\delta = 1/t$ in UCB, we
can solve for $N_2(T)$ when equality holds for the argument of $P$ in
Eq.~\ref{eq:full_inequality}. To simplify notation, we skip $T$ in the argument
of $N_2$:

\begin{equation}
	\begin{split}
		\sqrt{\frac{2}{N_2} \log \left(T\right)} & = \frac{\Delta_2}{2} \\
		N_{full} & = N_2 = \frac{8 \log(T)}{\Delta_2^2} 
	\end{split}
	\label{eq:exploration_full_point}
\end{equation}

Eq.~\ref{eq:exploration_full_point} carries physical meaning, as it implies that
if we explores for $N_{full}$ times, then the confidence bound will shrink below
to half of the suboptimality gap $\Delta_2$. In this case we can safely choose
$B_1$, and any more exploration is completely unnecessary. To put in a succinct
manner of speaking, by exploring $N_{full}$ times according to the above
equation, we fulfill the confidence bound of UCB.

Therefore, we write $N_2$ in Eq.~\ref{eq:exploration_full_point} as
\textit{Exploration Full Point} with $N_{full}$. Let $G_{full}$ denotes the
expected cumulative reward\footnote{For ease of analysis, we use reward instead
of regret for under exploration analysis. The same conclusion holds if regret is
used in this analysis.} at $N_{full}$:

\vspace{-1em}

\begin{equation}
	\begin{split}
		G_{full} = (T - N_{full}) \mu_1 + N_{full} \mu_2
	\end{split}
	\label{eq:gain_full}
\end{equation}

\subsubsection{Exploration Bargain Point} 
\label{sec:bargain_point}

The key question is whether we could stop exploring before $N_{full}$ and still
achieve better performance? The trade-off is that we have a smaller $N_2$ by
exploring less, and the probability of recognizing the non-optimal bandit as the
dominant choice $P(A_T = 2) = \delta$ will grow larger and become
non-negligible. Therefore, with a slight abuse of notation, we can write a lower
bound for expected cumulative reward as $G(N_2)$:

\vspace{-1em}
\begin{equation}
	\begin{split}
		G(N_2) & \ge \underbrace{\left( \left(T - N_2 \right) \mu_1 + N_2 \mu_2 \right)}_{\text{Correctly choose $B_1$ as the best one}} 
		 \overbrace{\left(1 - \delta \right)}^{\scaleto{\inf P(A_T=1)}{0.6em}} + \underbrace{\left( \left(T - N_2 \right) \mu_2 + N_2 \mu_1 \right)}_{\text{Mistakenly choose $B_2$ as the best one}} \overbrace{\delta}^{{\scaleto{\sup P(A_T=2)}{0.6em}}} \\
		\text{where } \delta & = e^{- N_2 \Delta^2_2 / 8} \text{ is from Eq.~\ref{eq:exploration_full_point}}
	\end{split}
	\label{eq:gain_bargain}
\end{equation}

Next, to ensure we can achieve better or at least the same performance as $N_{full}$, we write

\vspace{-1em}

\begin{equation}
	\begin{split}
		G(N_2) & \ge G_{full} \\ 
	\end{split}
	\label{eq:gN2_ge_gfull}
\end{equation}
 
Then, to find out the boundary condition where the same performance are
achieved, we can let $G_{full}$ in Eq.~\ref{eq:gain_full} to equal the right
hand side of the Eq.~\ref{eq:gain_bargain}.

\vspace{-1.0em}

\begin{equation}
	\begin{split}
		G_{full} & = \left( \left(T - N_2 \right) \mu_1 + N_2 \mu_2 \right)
		\left(1 - \delta \right) + \left( \left(T - N_2 \right) \mu_2 + N_2 \mu_1 \right) \delta
	\end{split}
	\label{eq:gfull_eq_right}
\end{equation}

We write the
solution of Eq.~\ref{eq:gfull_eq_right} as \textit{Exploration Bargain Point}
with $N_{bargain} = \underset{N_2}{\text{ solution}}  \{ \text{Equation~\ref{eq:gfull_eq_right}} \}$ because we achieve
at least the same expected reward by only exploring to the point of
$N_{bargain}$. We continue the
 discussion of the form of the solution in Appendix \ref{apn:ebp_more}.



\begin{figure}
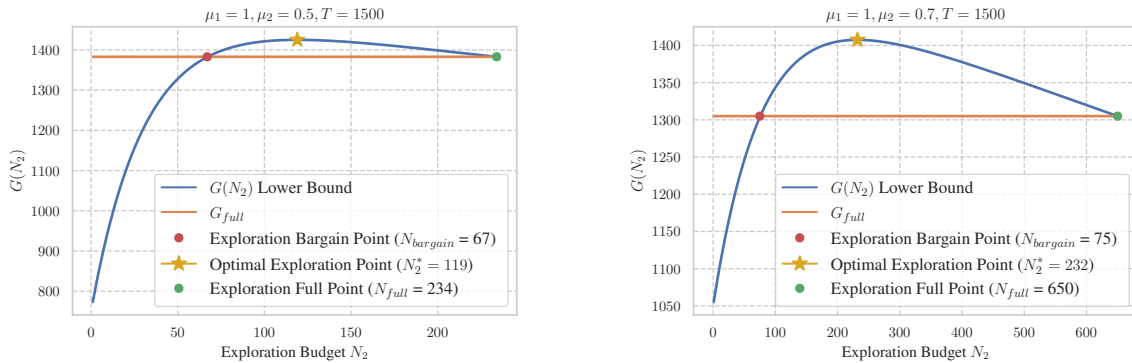

	\centering
	\begin{minipage}{.5\textwidth}
		\centering
		\scalebox{0.5}{
			\input{plots/pgf/ebp/g1.pgf}
		}
	\end{minipage}%
	\begin{minipage}{.5\textwidth}
		\centering
		\scalebox{0.5}{
			\input{plots/pgf/ebp/g2.pgf}
		}
	\end{minipage}
	\captionsetup{font={footnotesize}}
	\caption{Examples of the relationship between expected cumulative
		reward $G(N_2)$ and exploration budget $N_2$}
	\label{fig:example_ebp}
\end{figure}

\textbf{Determining $\mathbf{\gamma}$ with $\mathbf{N_{bargain}}$.} In Fig.
\ref{fig:example_ebp}, we visualize the relationship among $N_{full}$,
$N_{bargain}$, and $G(N_2)$ by examples. It is fascinating to see that the
optimal exploration point is located between $N_{bargain}$ and $N_{full}$
because of the concavity of the subgaussian reward distribution. Therefore, our
method can always perform better than UCB as long as $\gamma$ is set to $1 /
N_{bargain}$ in Eq.~\ref{eq:ucbdt}. It ensures that after exploring beyond
$N_{bargain}$ (\textcolor{DarkRed}{red} dot), the distance in Eq.~\ref{eq:ucbdt}
will become larger than 0, which makes the confidence bounds of all bandits in
Alg.~\ref{alg:ucbdt} smaller than that of UCB. By having smaller confidence
bounds, our policy will under-explore compared to UCB and stop before the
$N_{full}$ (\textcolor{DarkGreen}{green} dot)\footnote{A policy could stop
before $N_{bargain}$ if $\gamma$ is too high, while a $\gamma$ too low causes
the policy to approach $N_{full}$.}. Thus, the final $N_2$ will lie between
$N_{bargain}$ and $N_{full}$, where $G(N_2) > G_{full}$. Moreover, as shown by
Eq.~\ref{eq:transend} in Appendix~\ref{apn:ebp_more}, $N_{bargain}$ only depends
on suboptimality gap $\Delta_2$ and time horizon $T$, and does not rely on the
actual values of $\mu_1, \mu_2$. We argue that domain knowledge could be used to
estimate the difference between optimal bandit and other ones in practice and
thereby estimate $\gamma$.


\textbf{Existence of $\mathbf{N_{bargain}}$.} $N_{bargain}$ always exists and is
less than $N_{full}$ as
long as $\mu_1 > \mu_2$, and this can be proven by contradiction. Because of the
concavity of subgaussian distribution, the only possible case where this
condition is not satisfied is when $N_{bargain} = N_{full}$, in which the $G_{full}$
(\textcolor{YellowOrange}{orange} curve) becomes the tangent line of $G(N_2)$
(\textcolor{NavyBlue}{blue} curve). From Eq.~\ref{eq:gfull_eq_right}, we get

\vspace{-0.5em}
\small
\begin{equation}
	\begin{split}
		\left( \left(T - N_{full} \right) \mu_1 + N_{full} \mu_2 \right) = \left( \left(T - N_{full} \right) \mu_1 + N_{full} \mu_2 \right)
		\left(1 - \delta \right) + \left( \left(T - N_{full} \right) \mu_2 + N_{full} \mu_1 \right) \delta& 
	\end{split}
	\label{eq:existance_bargain}
\end{equation}
\normalsize
\vspace{-1em}

It follows directly that $\delta = 0$. Since $\delta = e^{- N_2 \Delta^2_2 / 8}$
from Eq.~\ref{eq:gain_bargain}, we get $N_2 = \inf$. Then, from
Eq.~\ref{eq:exploration_full_point}, $N_2 = \inf \rightarrow \Delta_2 = 0$.
However, we assume $\Delta_2 = \mu_1 - \mu_2 > 0$. So $N_{bargain}$ must always
exist.

\textbf{Implications.} Exploration Bargain Point describes exactly the
over-exploring nature of UCB. We could use $N_{bargain}$ as an anchor to
optimize the UCB method:

\begin{itemize}
	\item It allows UCB practitioners to early stop and prevent
	over-exploration. We summarize this policy as UCB-then-Commit in
	Appendix~\ref{apn:more_d}. In Table~\ref{tab:more_dist}, we see that UCB-then-Commit
	outperforms UCB in most experiments but is never as good as UCB-DT($\mu$).

	\item $N_{bargain}$ helps us understand when UCB-DT($\mu$) crosses into the optimal territory to
	get better rewards than UCB.
	
\end{itemize}

This insight is not available using traditional regret bound analysis. We
therefore believe that our analysis tool provides a novel and more intuitive
perspective on analyzing UCB-based methods.

\subsection{Regret Analysis}
\label{sec:analysis}


In addition to our analysis in Sec.~\ref{sec:under_exploration}, we offer a
finite time regret bound analysis for UCB-DT following standard practice. For
the sake of clarity of notation, we write $N_{i}(t)$ as $N_{i}(n)$. Suppose
$\mu_1 > \mu_2 > .. > \mu_k$, we can infer from Eq. \ref{eq:regret} that the
task of bounding $R_t$ can be translated as bounding
$\mathbb{E}\left[N_{i}(t)\right], i\in[k], i \neq 1$. The main idea is to divide the
analysis into two cases when we choose the suboptimal bandit over the optimal
one:


\begin{enumerate}[label=(\Alph*)]
	\item $\mu_1$ is under estimated, so the optimal bandit $B_1$ appears worse.
	\item $\mu_i, i\in[k], i\neq 1$ is over estimated, so the non-optimal bandit $B_i$ appears better.
\end{enumerate}

\vspace{-1em}
\small
\begin{equation}
	\hspace*{-20pt}
	\begin{aligned}
		N_{i}(n) = & \sum_{t=1}^{n} \mathbb{I}\left\{A_{t}=i\right\}                                                                                                                                                                                                                                                                                          \\
		\leq       & \underbrace{\sum_{t=1}^{n} \mathbb{I}\left\{\hat{\mu}_{1}(t)+\sqrt{\frac{2 \log (t) }{\widetilde{N}_1(t)}}  \leq \mu_{1}-\varepsilon\right\}}_{\text{(A)}}+ \underbrace{\sum_{t=1}^{n} \mathbb{I}\left\{\hat{\mu}_{i}(t)+\sqrt{\frac{2 \log t}{\widetilde{N}_i(t)}} \geq \mu_{1}-\varepsilon \text { and } A_{t}=i\right\}}_{\text{(B)}} \\
	\end{aligned}
\end{equation}
\normalsize

Based on Eq.~\ref{eq:Ntilde} we have $\widetilde{N}_i(t) \ge N_i(t)$, we can
therefore relax (B) by replacing $\widetilde{N}_i(t)$ with $N_i(t)$, which makes
it identical to the case of UCB and (B) can be bounded by $1 + \frac{2(\log t+\sqrt{\pi \log
t}+1)}{\left(\Delta_{i}-\varepsilon\right)^{2}}$ according to
\citep{lattimore2020bandit}. At the same time, while $\widetilde{N}_i(t) \le t$,
we cannot apply a similar procedure to (A) because directly replacing
$\widetilde{N}_i(t)$ with $t$ will make (A) grow much faster and violate the
\textit{sub-UCB} condition.


\begin{assumption}
	There exists a time step $\tau$, which ensures the policy chooses $A_1$ often
	enough:
	\vspace{-0.5em}
																																																																																																																																																																																																							
	$$
	\tau=\min \left\{t \leq T: \sup _{s \geq t}\left|\hat{\mu}_{1}(s)-\mu_{1}\right|<\varepsilon\right\}
	$$

	\vspace{-0.5em}
																																																																																																																																																																																																							
	\label{assumption:enough}
\end{assumption}

If we can leverage assumption \ref{assumption:enough}, then a simple observation
is that:

\begin{equation}
	\begin{aligned}
		\sum_{t=1}^{T} \mathbb{I}\left\{\hat{\mu}_{1}(t)+\sqrt{\frac{2 \log (t) }{\widetilde{N}_1(t)}}  \leq \mu_{1}-\varepsilon\right\} &   
		\le  \tau + \sum_{t=\tau+1}^T \mathbb{I}\{ \mu_1 - \hat{\mu}_1(t) \ge \varepsilon + \delta \}
	\end{aligned}
\end{equation}

It is straightforward to see that $P( \mu_1 - \hat{\mu}_1(t) \ge \varepsilon +
\delta \mid t > \tau) = 0$, where $\sqrt{2\log(t) / t}$ is replaced with $\delta
> 0$. Therefore, (A) can be bounded by $\tau$. Furthermore, if we let
bandit reward distribution to be gaussian, then by the concentration Lemma 1 in
\citep{lattimore2018refining}, $\mathbb{E}[\tau] \le 1+9 / \varepsilon^{2}$. Thus
the regret of UCB-DT satisfies the \textit{sub-UCB} requirement.

In the case of UCB-DT($\mu$), we can satisfy the assumption
\ref{assumption:enough} as long as $\gamma$ is small enough, such as $1 /
N_{bargain}$ according to Sec.~\ref{sec:bargain_point}. In fact, UCB-DT
can represent a family of policies through parameterization of $d$, and the
regret bound here only describes the boundary behavior as it approaches the standard UCB.
Based on our previous analysis,
there exists a parameter space of $d$ which can be used to customize the policy
to achieve better
exploration-exploitation trade-off. Therefore, instead of diving into a detailed
regret expression of any specific form, we provide a general analysis without
explicitly including this degree of freedom.


\section{Numerical Experiments}
\label{sec:exp}

We compare UCB-DT to UCB \citep{auer2002finite}, UCB$\dagger$
\citep{lattimore2018refining}, UCBV-Tune \citep{audibert2009exploration}, KL-UCB
\citep{garivier2011kl, cappe2013kullback} and KL-UCB++
\citep{menard2017minimax}. Among these methods, our main comparisons are done
for UCB, UCB$\dagger$ and UCBV-Tune. UCB-DT can also be
extended to support variance adaption. The remaining two methods, KL-UCB and
KL-UCB++, which deliver excellent performance, are \textit{non-anytime}
policies. Because of these structural differences, we do not regard them as our
main comparisons, however we still keep them as important references.

\begin{table}[h]
	\centering
	\begin{tabular}{lrrrrrr}
		\hline
		Experiment      & KL-UCB & KL-UCB++        & UCBV-Tune & UCB$\dagger$ & UCB     & UCB-DT($\mu$)  \\
		\hline
		B5 *            & 41.73  & \textbf{38.03}  & 98.6      & 129.01       & 251.28  & 70.95          \\
		B20 *           & 168.54 & \textbf{129.5}  & 415.23    & 410.5        & 939.99  & 383.82         \\
		B(0.02, 0.01) * & 28.34  & 22.39           & 64.99     & 131.99       & 249.17  & \textbf{21.6}  \\
		B(0.9, 0.88) *  & 38.44  & 33.78           & 49.6      & 73.9         & 119.91  & \textbf{19.19} \\
		N5 *            & 194.82 & 109.75          & 119.97    & 88.3         & 142.21  & \textbf{83.65} \\
		N20             & 539.91 & \textbf{410.25} & 840.26    & 560.47       & 1014.27 & 640.52         \\
		\hline
	\end{tabular}
	\vspace*{-2pt}
	\captionsetup{font={footnotesize}}
	\caption{Expected regret of different policies at $T=20000$ in each experiment. For each
		experiment we mark bold text for the best result and "*" if UCB-DT
		outperforms UCB, UCB$\dagger$ and UCBV-Tune.}
	\label{tab:regret}
\end{table}

For thorough evaluation, we design 6 experiments as below

\begin{description}[leftmargin=!,labelwidth=\widthof{\bfseries B(0.02, 0.01)}]
	\item[B5] Bernoulli reward, 5 bandits with expected rewards 0.9, 0.8, 0.7, 0.2,
	0.5. This experiment is modified from \citep{garivier2011kl} by adding more bandits.
	\item[B20] Bernoulli reward with many bandits, 20 bandits with expected rewards 0.9, 0.85, 0.8, 0.8, 0.7,
	0.65, 0.6, 0.6, 0.55, 0.5,
	0.4, 0.4, 0.35, 0.3, 0.3,
	0.25, 0.2, 0.15, 0.1, 0.05.
	\item[B(0.02, 0.01)] Bernoulli reward with low means, 3 bandits with expected rewards 0.05,
	0.02, 0.01. This experiment is borrowed
	identically from \citep{garivier2011kl}.
	\item[B(0.9, 0.88)] Bernoulli reward with close means, 2 bandits with expected rewards 0.9,
	0.88. 
	\item[N5] Gaussian reward, 5 bandits with unit variance and expected rewards
	1, 0.8, 0.5, 0.3, -0.2.
	\item[N20] Gaussian reward with many bandits, 20 bandits with unit variance
	and expected rewards 0, -0.03, -0.03, -0.07, -0.07, -0.07, -0.15, -0.15,
	-0.15, -0.5, -0.5, -0.5, -0.5, -0.5, -0.5, -0.5, -0.5, -0.5, -1, -1. This
	experiment is identical to the one	used in \citep{lattimore2018refining}.
\end{description}

We set $T$ to 20000 and run 2000 simulations for each method on every
experiment, and we set $\gamma$ to 0.02 in all experiments. For other methods,
we use implementations and default parameters from \citep{SMPyBandits}. We
summarize cumulative regrets in table \ref{tab:regret} and Fig. \ref{fig:exp}.

From our simulation results, we can see that UCB-DT is always better than UCB
and outperforms UCB$\dagger$, UCBV-Tune in first 5 experiments. The underlying
reason for the under performance of UCB-DT as compared to UCB$\dagger$ in N20
can be attributed to the high variance of the bandit reward distribution, which
makes the estimation of distance unstable. This performance degradation can be
mitigated by using a different distance as discussed in Appendix
\ref{apn:more_d}.


\section{Conclusion}

By leveraging bandit distance, we create a policy called UCB-DT, which is
simple, extensible and performant. Using our proposed framework, we propose the
concept of \textit{Exploration Bargain Point} to provide a new perspective on
analyzing performance of UCB-based methods. Admittedly, this work bears its own
limitations. We do not dive deeper in our regret bound analysis to describe the
relationship between convergence behavior and possible properties of $d$. We
only study the \textit{Exploration Bargain Point} in the context of our method
and standard UCB and have not applied it to other UCB-based policies. However,
we believe that these issues could be addressed in future work, and it may
be promising to adopt the idea in this paper to more general settings like
adversarial bandit and reinforcement learning.

\bibliographystyle{unsrtnat}









\clearpage

\appendix

\section{Regret Curves}
\label{apn:regret_curve}

\begin{figure}[H]
	\centering
	\scalebox{0.28}{
		\input{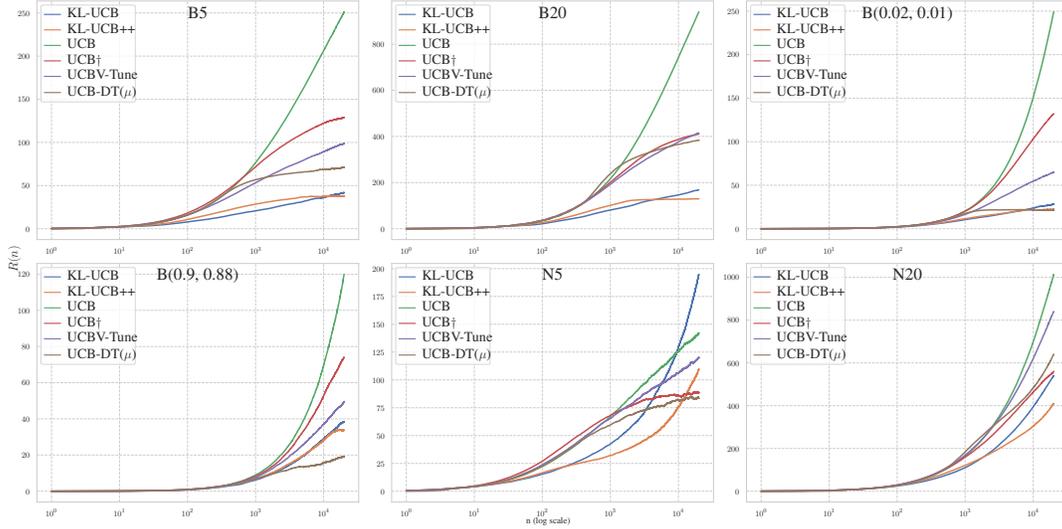}
	}
	\caption{Regret of different policies as a function of time (log scale) in experiments from Sec.~\ref{sec:exp}.}
	\label{fig:exp}
\end{figure}

\section{More Bandit Distances}
\label{apn:more_d}

We experiment the following two additional distances:

\begin{description}
	\item [UCB-then-Commit]
	$$
	d_t(i, j) = \begin{dcases}
	0 & N_i(t) \leq \floor{1 / \gamma} \\
	1 & N_i(t) > \floor{1 / \gamma}
	\end{dcases}
	$$
	\item [UCB-DT($\mu$ margin)] $d_t(i, j) = \left(|\hat{\mu}_i(t) -
	\hat{\mu}_j(t)| - m \right) ^{1 / \floor{\gamma N_i(t)}}$
\end{description}


We name the first strategy as UCB-then-Commit, which enables the transition from
exploration to exploitation occur based on $N_{bargain}$, thereby allowing
UCB practitioners to early stop and prevent over exploration. We also introduce a
second strategy called UCB-DT($\mu$ margin), which reduces the distance
$|\hat{\mu}_i(t) - \hat{\mu}_j(t)|$ by a margin. This distance reduction
encourages the policy to explore more among similar bandits and allows us to
better handle noisy environments with high variance and low mean. We summarize
their expected regrets in Table \ref{tab:more_dist}.

\begin{table}[H]
	\centering
	\begin{tabular}{lrrrr}
		\hline
		Experiment    & UCB     & UCB-DT($\mu$) & UCB-then-Commit &        
		UCB-DT($\mu$ margin)                                               \\
		\hline
		B5            & 251.28  & 70.95         & 76.11           & 73.49  \\
		B20           & 939.99  & 383.82        & 436.36          & 415.68 \\
		B(0.02, 0.01) & 249.17  & 21.6          & 58              & 243.04 \\
		B(0.9, 0.88)  & 119.91  & 19.19         & 105.7           & 110.01 \\
		N5            & 142.21  & 83.65         & 185.44          & 108.11 \\
		N20           & 1014.27 & 640.52        & 778.72          & 628.14 \\
		\hline
	\end{tabular}
	\vspace*{2pt}
	\caption{Expected regret of UCB and UCB-DT on more distances in each experiment.
		$\gamma$ is set to 0.02 in all experiments, $m$ is set to 0.05.}
	\label{tab:more_dist}
\end{table}

It is interesting to see that UCB-then-Commit generally outperforms UCB by a
large margin, which indicates the benefit of under exploration as in Sec.
\ref{sec:under_exploration}. UCB-DT($\mu$ margin) appears to be more robust than
UCB-DT($\mu$) in noisy environments by performing slightly better in N20, where
the mean is much smaller than variance compared to other experiments.


\section{Formulation of Standard UCB}
\label{apn:ucb}

According to \citep[Chapter~7]{lattimore2020bandit}, UCB is derived from
Hoeffding's inequality on sum of subgaussian variables. Let $X_{1}, X_{2},
\ldots, X_{n}$ be independent and L1-subgaussian random variables with zero mean
and $\hat{\mu}=\sum_{t=1}^{n} X_{t} / n$, then

\begin{equation}
	\mathbb{P}(\hat{\mu} \geq \varepsilon) \leq \exp \left(-n \varepsilon^{2} / 2\right)
	\label{eq:concen}
\end{equation}

Replacing $\exp \left(-n \varepsilon^{2} / 2\right)$ with $\delta$ then we get

\begin{equation}
	\mathbb{P}\left(|\hat{\mu}| \geq \sqrt{\frac{2}{n} \log \left(\frac{1}{\delta}\right)} \right) \leq \delta
	\label{eq:ucb_dawn}
\end{equation}

If we use $1/t$ for $\delta$, we arrive at the formulation of UCB.

\section{Optimal Exploration and Exploration Bargain Point}
\label{apn:ebp_more}

Using our analysis in Sec.~\ref{sec:under_exploration}, we gain a deeper
understanding as to why UCB is an "optimistic" policy. Furthermore, we can set
the derivative of the lower bound in Eq. \ref{eq:gain_bargain} on $N_2$ to 0 and
solve for the optimal bound, which represents the optimal
exploration-exploitation trade-off point. We calculate this result in Eq.
\ref{eq:N2star} with \citep{Mathematica}.

\begin{equation}
	\begin{split}
		&\left\{N_{2}^* \to \frac{\mu _1^2 T-2 \mu _2 \mu _1 T+\mu _2^2 T-16 W_{c_1}\left(\frac{1}{2} e^{\frac{\mu
			_1^2 T}{16}-\frac{1}{8} \mu _2 \mu _1 T+\frac{\mu _2^2 T}{16}+1}\right)+16}{2
		\left(\mu _2-\mu _1\right){}^2} \bigg| c_1 \in \mathbb{Z} \right\} \\
		& \text{where }  W \text{ denotes the Lambert $W$ function \citep{weisstein2002lambert}}
	\end{split}
	\label{eq:N2star}
\end{equation}

In the case of $N_{bargain}$, the exact solution cannot be found analytically
and we can only solve this numerically in Fig. \ref{fig:example_ebp}.  The
difficulty lies in Eq. \ref{eq:transend}, which is transcendental and has no
closed form expression for $N_2$ in this case.

\begin{equation}
	\begin{split}
		0 & = e^{-\frac{1}{16} \Delta ^2 _2 N_2} \left(2 N_2-T\right) - N_2+\frac{8 \log
			(T)}{\Delta _2^2} \\
		\text{the} & \text{ solution of this equation for $N_2$ is \textit{Exploration Bargain Point}}
	\end{split}
	\label{eq:transend}
\end{equation}

There are interesting insights which can be derived from these equations. For
example, based on Eq.~\ref{eq:N2star}, the optimal exploration point $N_{2}^*$
may not be unique. It will be interesting to study the relationship between the
solutions and determine which ones have practical relevance in future work.

\end{document}